\documentclass{article}

\PassOptionsToPackage{numbers, sort}{natbib}

\usepackage[preprint]{neurips_2021}

\usepackage[utf8]{inputenc} 
\usepackage[T1]{fontenc}    
\usepackage[hidelinks]{hyperref}       
\usepackage{url}            
\usepackage{booktabs}       
\usepackage{amsfonts}       
\usepackage{nicefrac}       
\usepackage{microtype}      
\usepackage{xcolor}         
\usepackage{algorithm}
\usepackage{algpseudocode}
\usepackage{amsmath}
\usepackage{graphicx}
\usepackage{subcaption}
\algnewcommand\algorithmicforeach{\textbf{for each}}
\algdef{S}[FOR]{ForEach}[1]{\algorithmicforeach\ #1\ \algorithmicdo}

\title{SEEN: Sharpening Explanations for Graph Neural Networks using Explanations from Neighborhoods}

\author{
  Hyeoncheol Cho \\
  Samsung SDS \\
  Seoul, Republic of Korea \\
  \texttt{hcheol35.cho@samsung.com} \\
  \And
  Youngrock Oh\thanks{This work was done when the author worked at Samsung SDS.} \\
  Mobilint \\
  Seoul, Republic of Korea \\
  \texttt{yrock.oh@gmail.com} \\
  \And
  Eunjoo Jeon \\
  Samsung SDS \\
  Seoul, Republic of Korea \\
  \texttt{ej85.jeon@samsung.com} \\
}

\begin{document}

\maketitle

\begin{abstract}
  Explaining the foundations for predictions obtained from graph neural networks (GNNs) is critical for credible use of GNN models for real-world problems.
  Owing to the rapid growth of GNN applications, recent progress in explaining predictions from GNNs, such as sensitivity analysis, perturbation methods, and attribution methods, showed great opportunities and possibilities for explaining GNN predictions.
  In this study, we propose a method to improve the explanation quality of node classification tasks that can be applied in a post hoc manner through aggregation of auxiliary explanations from important neighboring nodes, named $\textsc{Seen}$.
  Applying $\textsc{Seen}$ does not require modification of a graph and can be used with diverse explainability techniques due to its independent mechanism.
  Experiments on matching motif-participating nodes from a given graph show great improvement in explanation accuracy of up to 12.71\% and demonstrate the correlation between the auxiliary explanations and the enhanced explanation accuracy through leveraging their contributions.
  $\textsc{Seen}$ provides a simple but effective method to enhance the explanation quality of GNN model outputs, and this method is applicable in combination with most explainability techniques.
\end{abstract}

\section{Introduction} \label{introduction}
Learning and extracting information from graph structures are considered important but challenging, due to the difficulty of modeling relational information, even extending to long-range interactions among nodes~\cite{graphsage,gcn,xu2019spatio}.
Graph neural networks (GNNs), which are specially designed deep neural network for learning topologies and features from graphs, have revolutionized the field of machine learning on graph-structured data and achieved state-of-the-art performances~\cite{bruna,defferrard,gcn,gat}.
Recent progression on GNN architecture represented by a message-passing scheme, which recursively generates, aggregates, and updates node representations based on a local connectivity with neighborhoods, have generalized existing architectures and extended the applicability of GNNs to complex graph-structured problems~\cite{mpnn,graphsage,gin}.

Understanding why such decisions are made by GNNs improves the transparency of the models, helps to identify the failure modes and provides hints to revise the models.
Additionally, providing human-understandable explanations for GNN predictions is highly important for reliability and trustworthiness of GNNs, which are essential for critical applications requiring credible predictions, such as medical uses.
However, compared to the rapidly growing success of GNNs on graph-related tasks, explainability on GNN predictions has been less explored~\cite{gnngradcam,gnn_xai_taxonomy}.
There have been several successes on transferring explainability techniques developed for explaining convolutional neural networks (CNNs) to GNNs with minor modifications~\cite{gnngradcam, guidedbp_LRP_gnn}.
Recently, graph-oriented explainability methods have also been proposed to explain a prediction by extracting essential subgraphs from the input without changing the prediction~\cite{gnnexplainer,pgexplainer,graphmask,pgmexplainer}.
Generally, these methods provide explanations in the form of contribution scores for each component in the input graph for a given decision of a GNN.
Although those methods are designed to highlight important nodes or edges of the input graph for the target decision, their explanations often require additional models to be trained for generating graph masks~\cite{gnnexplainer,pgexplainer}.

We propose a method named Sharpening Explanations for graph neural networks using Explanations from Neighborhoods, abbreviated as $\textsc{Seen}$.
Given a prediction to be explained and an explainability method, $\textsc{Seen}$ enhances the target explanation by aggregating the auxiliary explanations from the assistant nodes near the target node.
Because graphs are used to represent relations between two nodes, we can assume that there is a strong correlation between the explanation of the target node and the explanations for its neighbors.
Specifically, we conjecture that given a pair of nodes, if the first node has a significant influence on the prediction of the second node, the explanation of the prediction of the former would be positively correlated to the target explanation to some extent.
In this regard, $\textsc{Seen}$ aggregates the auxiliary explanations by determining their importance weights based on the contribution scores of the corresponding nodes in the target explanation.
Since acquiring auxiliary explanations does not require alteration on neither the input graph nor the trained model, $\textsc{Seen}$ is capable of providing sharpened but still intact explanations based on the original model and data.
Moreover, it is worth noting that $\textsc{Seen}$ can be applied in a post-hoc manner with diverse explainability techniques on node classification tasks due to its independence in the way that individual explanations are generated.
We evaluated $\textsc{Seen}$ on widely used synthetic datasets for measuring explanation accuracy on node classification.
Our qualitative and quantitative evaluations demonstrated that applying $\textsc{Seen}$ can significantly improve explanation accuracy compared to the original evaluations.

\section{Related work}
\paragraph{Graph neural networks}
The foundation of GNNs has been presented by \citeauthor{bruna}~\cite{bruna} based on a spectral graph theory and has been expanded by numerous reports, such as \citeauthor{defferrard}~\cite{defferrard}, \citeauthor{gcn}~\cite{gcn}, and \citeauthor{duvenaud}~\cite{duvenaud}.
The message-passing scheme proposed by \citeauthor{mpnn}~\cite{mpnn}, which generalizes the GNN mechanism in terms of message, update, and readout functions, represents important progress on GNN formulations.
Most modern GNNs fall into the message-passing formulation, including graph convolutional network (GCN)~\cite{gcn}, graph attention network (GAT)~\cite{gat}, GraphSAGE~\cite{graphsage}, and graph isomorphism network (GIN)~\cite{gin}, which show outstanding performance on graph-structured data.
Advances in GNN models have made GNNs a favored model for various graph-related tasks, including node classification~\cite{gcn,graphsage,gat}, graph classification~\cite{duvenaud,gin,mpnn}, and link prediction~\cite{zhang2018link,kazemi2018simple}.
In our paper, we utilize the GCN architecture as the model system to be explained when evaluating $\textsc{Seen}$, based on its popularity in learning graph-structured data and expandability to diverse message-passing GNNs without losing generality.

\paragraph{Explainability methods for GNN}
There has been an increasing number of works that study explainability methods for deep neural networks, especially CNNs: gradient/feature-based methods~\cite{gradient, cam, gradcam, gradcam_plus}, perturbation-based methods~\cite{Zint_2017, Fong_2017, Fong_2019_ICCV}, and decomposition methods~\cite{lrp, deeplift}.
Inspired by these studies, the interpretability of GNNs has also been addressed by similar approaches~\cite{gnn_xai_taxonomy}.
Gradient/feature-based methods are proposed, including SA~\cite{guidedbp_LRP_gnn}, Guided BP~\cite{guidedbp_LRP_gnn}, CAM~\cite{gnngradcam}, and Grad-CAM~\cite{gnngradcam}.
Several perturbation-based methods are proposed including GNNExplainer~\cite{gnnexplainer}, PGExplainer~\cite{pgexplainer}, ZORRO~\cite{zorro}, GraphMask~\cite{graphmask}, and Causal Screening~\cite{causal}.
Decomposition-based methods are also applied to explain the deep graph neural networks, including Layerwise Relevance Propagation (LRP)~\cite{guidedbp_LRP_gnn}, Excitation BP~\cite{gnngradcam} and GNN-LRP~\cite{gnn_lrp}.

On the other hand, a few papers have proposed methods to enhance explanations of a given explainability method for CNNs~\cite{smoothgrad,evet}.
In general, explanation enhancement methods make copies of the input image with a small perturbation and incorporate the explanations for them to give a better explanation for the target prediction.
For example, SmoothGrad~\cite{smoothgrad} takes random samples in the neighborhood of the input and average the explanations of the samples.
EVET~\cite{evet} provides a visually clear explanation that takes into account geometric transformations of the input image.
$\textsc{Seen}$ is similar to them in the way that we incorporate the auxiliary explanations to sharpen the target explanation.
However, $\textsc{Seen}$ do not modify the input graph, instead it gathers neighboring nodes from the graph and integrates the explanations for them, whereas Smoothgrad and EVET modify the input image to obtain the auxiliary explanations.

\section{Preliminaries} \label{prelim}
\paragraph{Graph neural networks}
Define a graph $\mathcal{G}=\left(V,E\right)$ by a set containing $N$ nodes $V=\{v_1, ..., v_N\}$ and a set containing $M$ edges $E=\{e_1, ..., e_M\}$.
A GNN model $\Phi$ takes a graph $\mathcal{G}$ as an input and performs node-level, graph-level, or edge-level predictions through a series of message-passing layers and a pooling layer if required.
The input graph $\mathcal{G}$ can be presented by three matrices, an adjacency matrix $A \in \{ 0, 1 \}^{N \times N}$, a node feature matrix $X_v \in \mathbb{R}^{N \times D}$, and an optional edge feature matrix $X_e \in \mathbb{R}^{N \times N \times K}$ if edge features are provided, where $D$ and $K$ denote the number of features for nodes and the number of features for edges, respectively.

Each layer of GNNs based on the message-passing scheme can be divided into three steps: message generation, neighborhood aggregation, and representation update steps~\cite{mpnn}.
In the message generation step, the message function $\textsc{Message}$ takes the node representations $h_i$ and $h_j$ of edge $\left(v_i, v_j\right)$ and its edge representation $h_{ij}$ from previous layer and calculates message $m_{ij}=\textsc{Message}\left(h_i, h_j, h_{ij}\right)$.
The neighborhood aggregation step collects messages through the aggregate function $\textsc{Aggregate}$ from neighboring nodes $\mathcal{N}_i$ of $v_i$, generating aggregated message $m'_i=\textsc{Aggregate}\left(\{m_{ij} \mid v_j \in \mathcal{N}_i\}\right)$.
The representation update step merges aggregated message $m'_i$ with previous representation $h_i$ through the $\textsc{Update}$ function.
The updated representation $h'_i=\textsc{Update}\left(h_i, m'_i\right)$ becomes the node representation for node $v_i$ in the current layer and is propagated to further layers.

\paragraph{Explaining GNN predictions}
We focus on explaining a node classification model, which is the target scope of this paper.
Given a graph $\mathcal{G}$, a node classification model $\Phi:\mathcal{G}\to y$, and an target node $v_t$, an explainability technique estimates an explanation $S(v_t)$ for a model prediction $\Phi \left(G, v_t \right)$.
An explanation $S(v_t)$ can be either a set of node scores $S(v_t)=\{s_v \mid v\in V\}$ or a set of edge scores $S(v_t)=\{s_e \mid e\in E\}$, depending on the explainability methods.

\section{$\textsc{Seen}$: Sharpening Explanations using Explanations from Neighborhoods} \label{seen}
\begin{figure}[ht]
  \includegraphics[width=\textwidth]{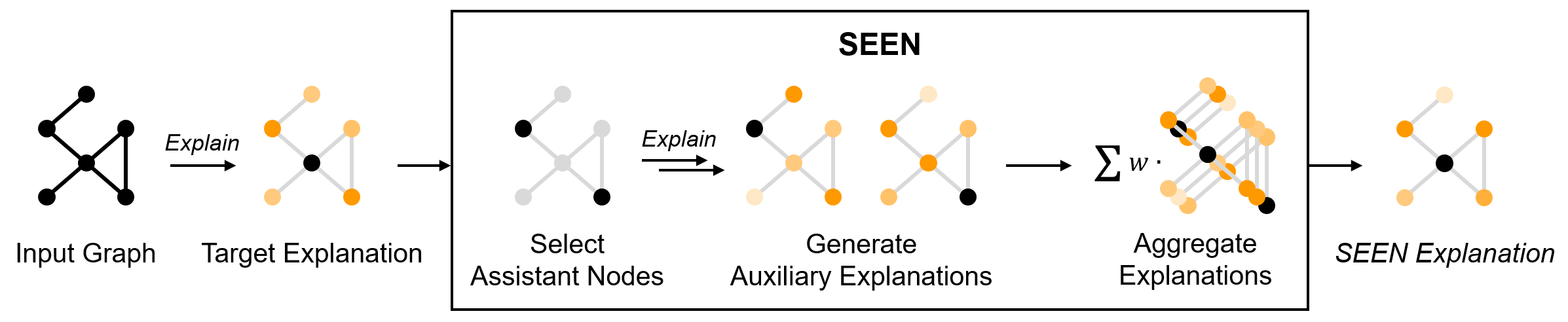}
  \caption{Schematic for sharpening the explanation with $\textsc{Seen}$ on node classification. With the input graph, trained model and explanation target, $\textsc{Seen}$ performs explanation sharpening through selecting assistant nodes, generating auxiliary explanations from the assistant nodes, and aggregating the target and auxiliary explanations.}
  \label{fig:scheme}
\end{figure}

In this section, we introduce our explanation sharpening method using neighborhood explanations for enhancing the explanation quality on node classification tasks.
Our method, $\textsc{Seen}$, accumulates auxiliary explanations from assistant nodes of a target node, without modifications on an input graph.
Given the explanation target node $v_t$, our method collects the set of assistant nodes $V_a = \{v_a \mid v_a \ne v_t, v_a\in V\}$ from the input graph $\mathcal{G}$ and performs auxiliary explanation for each assistant node $v_a$.
Obtained auxiliary explanations are aggregated with the original explanation from the target node, considering their importance on the target prediction.
It should be noted that $\textsc{Seen}$ is a post hoc process that is attachable to the original explainability techniques and that $\textsc{Seen}$ gathers additional information and updates the original explanation.
The overview for the $\textsc{Seen}$ is depicted in Figure~\ref{fig:scheme}.
We begin this section with the detailed process for the auxiliary explanation acquisition (Section~\ref{aux_explanation}) and then discuss importance-based explanation aggregation for sharpening the original explanation with the auxiliary explanations (Section~\ref{exp_aggregation}).
The entire process is presented in Algorithm~\ref{alg:seen}.
We discuss the motivation on our explanation sharpening method (Section~\ref{motivation}).

\subsection{Auxiliary explanation acquisition} \label{aux_explanation}
The first step of the auxiliary explanation acquisition is the selection of assistant nodes from which to collect explanations.
Selecting an appropriate set of assistant nodes $V_a$ is crucial for obtaining helpful auxiliary explanations.
To collect meaningful and supportive assistant nodes, we set a distance-based boundary on the assistant node pool.
Distance-based boundary methods measure the number of edges in the shortest path between a candidate node and a target node to exclude unnecessary nodes for explanation, which acts as a boundary for sampling assistant nodes from an input graph.
We filter the nodes outside of the $k$-hop neighborhood of the target node $v_t$ when $k$ message-passing layers exist in a model, instead of considering all nodes within the entire graph.
The nodes outside of the $k$-hop neighborhood have zero influences on the prediction for the target node, which is the main focus to be explained, and thus, is considered to be inappropriate for collecting auxiliary explanations for the target.

When the assistant nodes are prepared, the explainability technique is applied to generate explanations for the assistant nodes.
It should be noted that explaining the assistant nodes is not perfectly identical to explaining the target node.
Given the explainability technique $\textsc{Explain}$, target node $v_t$, and model prediction $y_t$, the logit $y^c$ with the most probable class label $c$, is utilized to explain the target node.
On the other hand, to explain the assistant node $v_a$, the logit $y_a^c$ with the same class label $c$ is employed, regardless of the predicted class for $v_a$:
\begin{align}
  S(v_t) &= \textsc{Explain} \left( \mathcal{G}, \,v_t \right) = \textsc{Explain} \left( \mathcal{G}, \,y_t^c \right), \quad
  y_t^c = \Phi \left(G, \,v_t \right)^c \label{eq:S_v_t} \\
  S(v_a) &= \textsc{Explain} \left( \mathcal{G}, \,v_a \right) = \textsc{Explain} \left( \mathcal{G}, \,y_a^c \right), \quad
  y_a^c = \Phi \left(G, \,v_a \right)^c \label{eq:S_v_a}
\end{align}
where $\textsc{Explain}$ is the explainability technique, $c$ is the true class label for $v_t$, and $\Phi(G, v)^c$ is the logit for model prediction on $v$ with class $c$.
Using the logit with the same class label for explanation allows sharing explanations across the nodes.

\subsection{Explanation aggregation} \label{exp_aggregation}
With the explanations $S(v_t)$ and $\{S(v_a) \mid v_a \in V_a\}$ that are obtained from the target node $v_t$ and assistant node $V_a$, respectively, a summarized explanation $\bar{S}(v_t)$ is calculated as a final explanation for the target node $v_t$.
When aggregating explanations, the choice of aggregation formula can be diverse.
Here, we hypothesize that the auxiliary explanation obtained from the important assistant node would be more influential and supportive for the target explanation.
In detail, we assign a high weight to the auxiliary explanation generated by the assistant node that had high importance in the original target explanation, whereas a low weight is assigned to the auxiliary explanation from the low-importance assistant node.
To model the importance-based weighted summation of auxiliary explanations, we design our aggregation formula to incorporate two coefficients, $\alpha$ and $\beta$, for modeling the significance of auxiliary explanations and exponentially decaying weight with respect to the importance ranking of assistant nodes.
It is possible to use an arbitrary decaying series to model the decaying weights.
We chose a simple, exponentially decaying series for our study to examine the efficacy of our system (Equation~\ref{eq:seen}).
\begin{equation} \label{eq:seen}
  \bar{S}(v_t) = S(v_t) + \alpha \sum_{r=1}^{\lvert V_a \rvert} \beta^{r-1} S(v^{(r)})
\end{equation}
where $\lvert V_a \rvert$ is the number of assistant nodes within the assistant node set, $v^{(r)}$ is the $r$th assistant node by the decreasing importance score from the target explanation $S(v_t)$, $\alpha$ is the weight coefficient for auxiliary explanations in range $\left[ 0, 1 \right]$, and $\beta$ is the decay coefficient in range $\left[ 0, 1 \right)$ for addressing auxiliary explanations with low importance scores.
The coefficients $\alpha$ and $\beta$ are shared within a graph in which identical values are applied to explain all nodes for a given dataset and explainability technique combination.

The equation can be simplified in particular combinations of $\alpha$ and $\beta$; for example, when $\alpha$ is set to 0, the equation disregards all auxiliary explanations similar to common explanation techniques.
If $\alpha > 0$ and $\beta \to 1$, the equation models equivalent handling of auxiliary explanations regardless of the importance of its origin node:
\begin{equation}
  \bar{S}(v_t) \approx S(v_t) + \alpha \!\!\sum_{v_a \in V_a}\!\! S(v_a)
\end{equation}

\begin{algorithm}
\caption{Shapening Explanations using $\textsc{Seen}$}
\label{alg:seen}
\begin{algorithmic}[1]
\Procedure{$\textsc{Seen}$}{Graph $\mathcal{G}=(V,E)$, target node $v_t$, explainability technique $\textsc{Explain}$, number of message-passing layers $k$, coefficients $\alpha$ and $\beta$}
  \State $S(v_t) \gets \Call{Explain}{\mathcal{G}, \,v_t}$
  \Comment{Eq. \eqref{eq:S_v_t}}

  \State $\text{Update}\, \bar{S}(v_t) \gets S(v_t)$

  \ForEach {$v_a \in V$}
    \If {$0 < \Call{Distance}{v_t, v_a} \leq k$}
    \Comment{Distance-based boundary}
      \State $r \gets \Call{Rank}{S(v_t), v_a}$
      \Comment{Rank $v_a$ by $s_a \in S(v_t)$}

      \State $S(v_a) \gets \Call{Explain}{\mathcal{G}, \,v_a}$
      \Comment{Eq. \eqref{eq:S_v_a}}

      \State $\text{Update}\, \bar{S}(v_t) \gets \bar{S}(v_t) + \alpha \beta^{r-1} S(v_a)$
      \Comment{Eq. \eqref{eq:seen}}
    \EndIf
	\EndFor
\EndProcedure
\end{algorithmic}
\end{algorithm}

\subsection{Motivation} \label{motivation}
The intuition behind the collection of auxiliary explanations of $\textsc{Seen}$ is to gather information across the receptive fields of GNNs and earn a (locally) shared explanation.
Each message-passing layer of a GNN transfers information from each node to their neighboring nodes, expanding the receptive field by one hop.
With a $k$-layered GNN, the receptive field corresponds to the $k$-hop neighborhood of the prediction target.
Because all nodes within the $k$-hop contribute to the prediction for a target node, a bounding assistant node pool is needed in the $k$-hop neighborhood.

Within the receptive field, $\textsc{Seen}$ aggregates auxiliary explanations in order of their node importance from the target explanation $S(v_t)$ to refine explanations through overlaying neighborhood explanations.
In terms of a community detection problem that predicts each node to certain classes, predicting the class for a node near the community boundary would suffer from ambiguity of the boundary criteria.
Neighborhoods of the node that partially share the criteria can help to sharpen the boundary to decide to which community the target belongs, due to the local proximity between the target node and neighborhoods.
Accumulating and overlaying the boundary criteria within the local neighborhoods would refine the original criteria and can be extended to sharpening explanations for the predictions.
Based on this idea, we conjecture that the explanation could be sharpened by accumulation of auxiliary explanations from neighboring nodes that have a nonzero contribution to the target prediction.

\section{Experiments} \label{experiments}
To assess the effectiveness of our conjecture, we conducted a series of experiments on explaining predictions from graph neural networks in node classification tasks.
First, we describe datasets (Section~\ref{datasets}) and the model employed for training and explaining graph neural networks (Section~\ref{model}).
Second we present explainability techniques to be examined (Section~\ref{explainability}) and evaluation criteria (Section~\ref{evaluation}) for measuring the performance of $\textsc{Seen}$.
Through qualitative and quantitative analyses, we demonstrate that our method can improve the explanation accuracy to a maximum of 12.71\% in the best performing case, without losing accuracy in the least-performing cases.
Screening the aggregation coefficients $\alpha$ and $\beta$ shows that our conjecture, which seeks sharpening explanations through accumulation of explanations from neighborhoods, is valid.
Moreover, the screening process also reveals that a certain trend exists for coefficient-performance relation and assigning appropriate values is highly important for maximizing the sharpening effect of $\textsc{Seen}$.

\subsection{Datasets} \label{datasets}
We utilize four widely used public synthetic datasets for evaluating explainability techniques on node classification tasks constructed by \citeauthor{gnnexplainer} \cite{gnnexplainer}: BA-Shapes, BA-Community, Tree-Cycles, and Tree-Grid.
All datasets pursue the classification of each node into their role in attached motifs, including not-participating positions.
The BA-Shapes dataset contains Barabasi-Albert (BA) graphs decorated with house-structured motifs in random positions.
Nodes are numbered 0 to 3 for not-participating, located on the top, middle, and bottom of the motif.
The BA-Community dataset is a set of graphs made by a combination of two graphs from the BA-Shapes dataset with doubled node classes.
Each node in the BA-Shapes subgraph is given with the node feature based on its community membership.
The Tree-Cycles dataset contains an 8-level tree with randomly attached hexagonal cycle motifs.
The task is to binary classify each node regardless of whether it participates in cycle motifs.
The Tree-Grid dataset is a replacement of the cycle motifs into 3-by-3 grid motifs from the Tree-Cycles dataset.
No node features are provided for any datasets, except for the BA-Community dataset. Data splits for each dataset are taken from the code by \citeauthor{pgexplainer} \cite{pgexplainer}, which divides the dataset into 80\%, 10\%, and 10\% portions for training, validation, and testing, respectively.

\subsection{Model} \label{model}
All of experiments are conducted with a GCN model~\cite{gcn} for node classification with three message-passing layers.
The GCN model is trained with cross-entropy loss for ten different random seed values to generate ten individually trained models.
The trained models are then frozen and shared by all explanation experiments.
Detailed architecture, training hyperparameters, and averaged model accuracies for the GCN model are provided in Appendix.

\subsection{Explainability methods} \label{explainability}
To evaluate the explanation sharpening performance of $\textsc{Seen}$, we adopt the following three GNN explainability techniques, which are compatible with explaining node classification tasks: sensitivity analysis (SA)~\cite{gradient, guidedbp_LRP_gnn, gnngradcam}, Grad*Input~\cite{deeplift, attribution}, and GradCAM~\cite{gradcam, gnngradcam, attribution}.
Explainability methods based on gradients and features are particularly well-suited for $\textsc{Seen}$ due to their fast and no-training-required characteristics. However, we would like to emphasize that $\textsc{Seen}$ is not associated with specific techniques for the enhancement target.
When calculating gradients was required for explaining assistant nodes, we measured te gradient of the logit that belongs to the same class as the target node is predicted to be, as previously mentioned in Equation~\ref{eq:S_v_a} in Section~\ref{aux_explanation}.

\paragraph{Sensitivity Analysis}
We utilize the basic SA method~\cite{gradient}, which calculates the gradient of the prediction with respect to the features and generates scores by taking the absolute value of the gradient.
SA measures the influence of each input value on the final prediction through the gradient and assumes that higher absolute gradient values indicate the higher importance.
Due to its simplicity, transferring SA to GNNs can be easily done by calculating the gradient with respect to the node features instead of pixels~\cite{guidedbp_LRP_gnn, gnngradcam}.
It is possible to apply SA on explaining predictions with scoring both nodes and feature elements, we focus on scoring each node in this paper.

\paragraph{Grad*Input}
Similar to the SA on GNNs, Grad*Input~\cite{deeplift, attribution} on GNNs calculates the explanation through element-wise multiplication of node features and their gradient over the prediction and the consequent summation over the feature dimension to generate node scores.
Compared to the SA, the gradient from Grad*Input is directly multiplied to the input, and the absolute value is obtained after the multiplication.

\paragraph{GradCAM}
GradCAM~\cite{gradcam} generalizes the class activation map (CAM) method~\cite{cam}, which requires a global average pooling layer in the model architecture, with gradients from each layer.
Node representations from intermediate message-passing layers are element-wisely multiplied with their gradients over prediction and summed into a single score value.
A detailed formulation for applying GradCAM on GNNs varies from reported papers~\cite{gnngradcam, attribution}, such as the selection of the target layer, averaging the gradients node-wisely or taking the absolute value; we take an approach similar to the GradCAM(all) method applied by \citeauthor{attribution} \cite{attribution}, which multiplies the gradient and features element-wisely and averages the scores from each layer. Absolute values are obtained at the final stage of the explanation.

\subsection{Evaluation} \label{evaluation}
We perform experiments with all combinations of the aforementioned four datasets and three explainability techniques to quantify the explanation sharpening ability of $\textsc{Seen}$.
Since the ground-truth explanation (the binary labels whether each node is motif-participating) is available for the datasets due to their synthetic nature, we measure the area under the receiver operation characteristic curve (AUC-ROC) between the ground-truth and the obtained explanations, similar to \citeauthor{gnnexplainer}~\cite{gnnexplainer} and \citeauthor{attribution}~\cite{attribution}.
Based on the measure, we calculate the difference in the explanation accuracy with or without $\textsc{Seen}$ on each combination.

It should be noted that a modification is made on the test set of the datasets when compared to the original protocol by \citeauthor{gnnexplainer}
Originally, an evaluation is conducted only for the set of selected nodes, one node per one motif.
Instead, we perform an evaluation for all nodes participating in the motifs instead of using the selected node set.
For example, all five nodes that construct a house-structured motif in the BA datasets are accounted for in the evaluation target, rather than utilizing only the left top node of each house-structured motif.

\subsection{Results}
\paragraph{Evaluation on explanation sharpening}
Table~\ref{table:accuracy} shows the change in the explanation accuracy when $\textsc{Seen}$ is applied to different pairs of datasets and explainability techniques.
The significance of the difference is analyzed with a one-sided paired t-test or a one-sided Wilcoxon signed-rank test based on the normality test results.
Statistical analyses show that $\textsc{Seen}$ significantly increases the accuracy on providing motif-participating nodes as an explanation for the predictions, to a maximum of 12.71\% on the BA-Community dataset.
In the least-performing cases, the measured explanation accuracy is equivalent to the original explanation when statistically analyzed.
An extended table with the standard deviations, p-values, and aggregation coefficients is included in Appendix.
These results indicate that $\textsc{Seen}$ successfully generates a sharpened explanation with higher accuracies through collecting supportive auxiliary explanations around the target node.

\begin{table}[ht]
  \caption{Explanation accuracies (AUC-ROC) before and after applying $\textsc{Seen}$. The best performing accuracies among the combinations of $\alpha$ and $\beta$ are listed for each dataset and explainability technique. Differences that are not significant by statistical analyses are marked as n.s.}
  \vspace*{1mm}
  \label{table:accuracy}
  \centering
  \begin{tabular}{lcccc}
  \toprule
  \textbf{Explainability Methods}
  & \textbf{BA-Shapes}          & \textbf{BA-Community}
  & \textbf{Tree-Cycles}        & \textbf{Tree-Grid} \\
  \midrule
  SA
  & 0.935            & 0.625            & 0.886            & 0.814            \\
  SA + $\textsc{Seen}$
  & 0.938            & 0.637            & 0.900            & 0.866            \\
  \textit{Improvement}
  & \textit{n.s.}    & 1.85\%          & \textit{n.s.}    & \textit{n.s.}    \\
  \midrule
  Grad*Input
  & 0.925            & 0.614            & 0.894            & 0.722            \\
  Grad*Input + $\textsc{Seen}$
  & 0.936            & 0.618            & 0.934            & 0.800            \\
  \textit{Improvement}
  & 1.18\%           & 0.52\%           & 4.46\%           & 10.77\%          \\
  \midrule
  GradCAM
  & 0.904            & 0.573            & 0.750            & 0.757            \\
  GradCAM + $\textsc{Seen}$
  & 0.918            & 0.645            & 0.756            & 0.785            \\
  \textit{Improvement}
  & 1.62\%           & 12.71\%          & \textit{n.s.}    & \textit{n.s.}    \\
  \bottomrule
  \end{tabular}
\end{table}

\paragraph{Analysis on explanation aggregation coefficients}
Our aggregation design leverages the amount of auxiliary explanation to be incorporated with the target explanation through two coefficients, $\alpha$ and $\beta$.
To understand the performance trend with respect to the coefficients, we perform a grid scan over the coefficients by a 0.25 increment within $\left[0, 1\right]$ for $\alpha$ and $\left[0, 1\right)$ for $\beta$ and measure the explanation accuracy.
The $\beta=1$ configurations are additionally tested but excluded from the results for visual clarity due to the dramatic decrease in performance compared to the other configurations (data not shown).
The low performance for $\beta=1$ suggests that equivalently taking into account all auxiliary explanations is not beneficial for sharpening the explanation. Thus, assigning an appropriate series of decreasing weights when aggregating explanations is necessary for accuracy enhancement.

\begin{figure}[ht]
  \centering
  \begin{subfigure}[b]{0.3\textwidth}
    \includegraphics[trim={1.5cm 0 0 2cm}, clip, width=\textwidth]{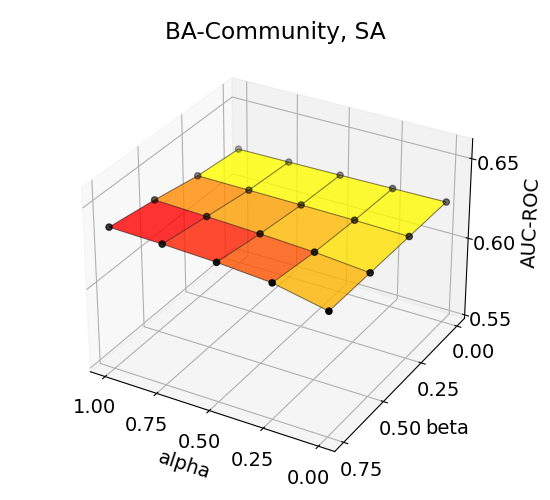}
    \caption{BA-Community, SA}
  \end{subfigure}
  \hfill
  \begin{subfigure}[b]{0.3\textwidth}
    \includegraphics[trim={1.5cm 0 0 2cm}, clip, width=\textwidth]{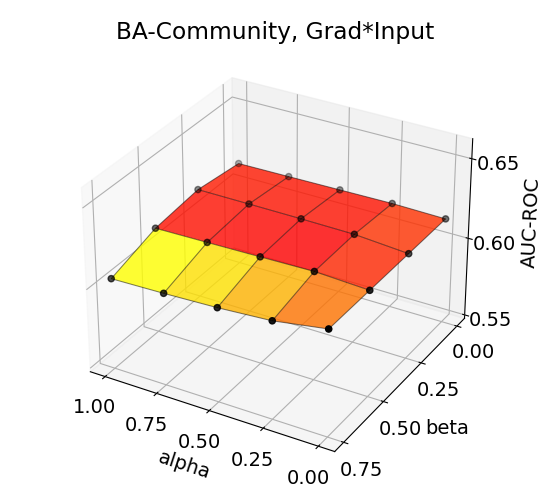}
    \caption{BA-Community, Grad*Input}
  \end{subfigure}
  \hfill
  \begin{subfigure}[b]{0.3\textwidth}
    \includegraphics[trim={1.5cm 0 0 2cm}, clip, width=\textwidth]{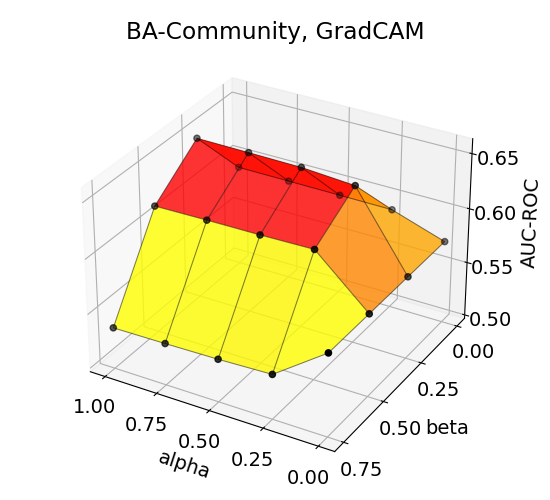}
    \caption{BA-Community, GradCAM}
  \end{subfigure}
  \newline

  \begin{subfigure}[b]{0.3\textwidth}
    \includegraphics[trim={1.5cm 0 0 2cm}, clip, width=\textwidth]{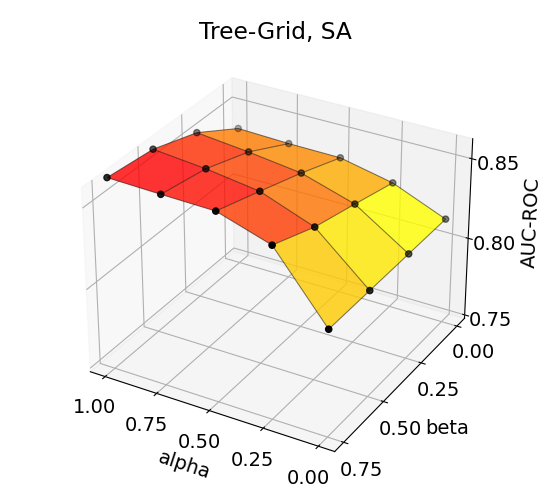}
    \caption{Tree-Grid, SA}
  \end{subfigure}
  \hfill
  \begin{subfigure}[b]{0.3\textwidth}
    \includegraphics[trim={1.5cm 0 0 2cm}, clip, width=\textwidth]{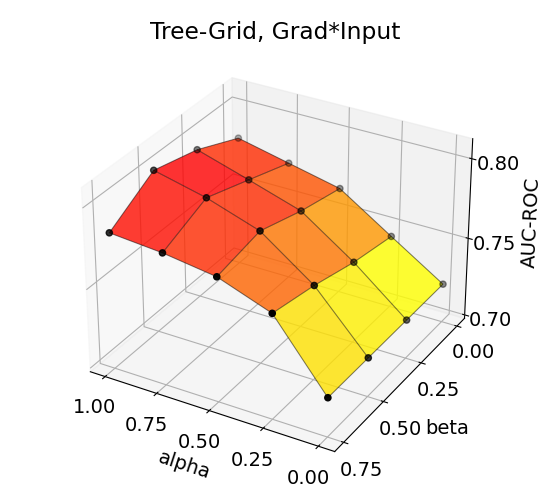}
    \caption{Tree-Grid, Grad*Input}
  \end{subfigure}
  \hfill
  \begin{subfigure}[b]{0.3\textwidth}
    \includegraphics[trim={1.5cm 0 0 2cm}, clip, width=\textwidth]{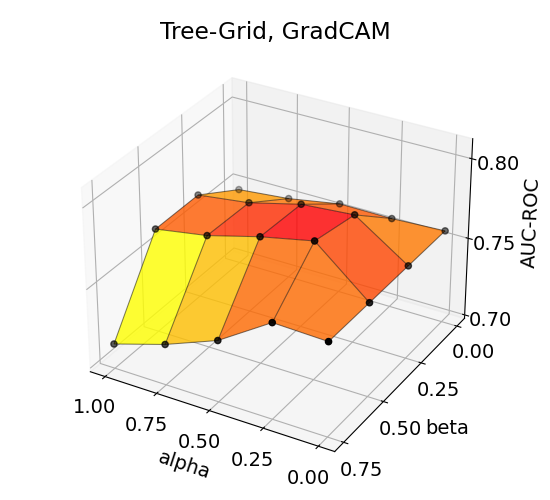}
    \caption{Tree-Grid, GradCAM}
  \end{subfigure}

  \caption{Effects of aggregation coefficients on explanation accuracy for BA-Community and Tree-Grid datasets. The results for the remaining combinations are listed in the Appendix.}
  \label{fig:coefficients}
\end{figure}

The accuracy trends for BA-Community and Tree-Grid datasets (Figure~\ref{fig:coefficients}) show that maximum performance is achieved in high $\alpha$ and medium-to-high $\beta$, indicating the helpfulness of auxiliary explanations toward the target explanation.
The performance trends for all combinations are depicted in Appendix.
Moreover, a similar performance trend is observed in other datasets, proving that our hypothesis on sharpening an explanation with importance-ranked neighborhood explanations.
Note that accuracies change even up to 0.1 in the AUC-ROC upon the choice of $\alpha$ and $\beta$. We recommend conducting a parameter scan to obtain a higher performance increase with $\textsc{Seen}$.

An interesting observation is that GradCAM shows distinctive performance patterns and that the maximum performance is obtained in low $\alpha$ and medium $\beta$, compared to other methods that generally reach their maximum at high $\alpha$ and high $\beta$.
We assume that the difference may be attributable to the difference in handling explanation-hop relations; GradCAM collects explanations from each and all message-passing layers/hops of GNN, whereas SA and Grad*Input generate explanation within the input layer, which is basically 0-hop.
The performances for GradCAM with $\alpha=0.25$ are still shown to be higher than those with $\alpha=0$, suggesting that auxiliary explanations are supportive in all cases.

\section{Limitations} \label{limitations}
Before finishing, we describe limitations of our work in three aspects: applicable task, dataset, and explainability method.
Fundamental mechanism of $\textsc{Seen}$, which acquires additional explanations from assistant nodes, limits the applicable scope of $\textsc{Seen}$ to node-level prediction tasks.
The graph-level prediction tasks are not applicable due to the inability of collecting multiple explanations from neighborhoods.
Next, our experiments are conducted with synthetic datasets.
It is worth noting that "correct" explanations on graphs are considered to be more ambiguous compared to images, that synthetic dataset with ground-truth explanation is necessary for evalaution.
Still, applying $\textsc{Seen}$ on real-world data may have different aspects and trends from the empirical results presented in this paper.
Finally, the employed explainability methods in our paper are gradient/feature-based methods.
Recent methods involving graph masks are compatible with $\textsc{Seen}$, but investigation on assigning importance-based weights is necessary due to their edge-based scoring mechanism.

\section{Conclusion}
In this paper, we propose $\textsc{Seen}$, an explanation sharpening method using auxiliary explanations obtained from neighboring nodes.
Through aggregation of auxiliary explanations with importance-based weights, $\textsc{Seen}$ generates a neighborhood-shared explanation without a modification of the data or an extra model training for explanation enhancement.
Our experiments show that our simple method can significantly increase the explanation accuracy, in conjunction with the open choice for the explainability technique.
Moreover, accuracy trends of aggregation coefficients confirm that our strategy to assign high weights on important nodes is suitable for sharpening explanations by overlaying auxiliary explanations.
We expect that our approach would improve the reliability of GNN predictions and expand the scope of the explainability technique to various GNN applications.

\section*{Broader Impact}
Explaining predictions from graph neural networks are becoming more important, with the expanding use of the GNN model on graph-structured data.
Explainability techniques help users to obtain hints on why such a prediction is made and allow them to revise the model.
This paper proposed an explanation sharpening method that utilizes auxiliary explanations from neighborhoods and demonstrated the efficacy on increasing the explanation accuracy.
Moreover, the post hoc attachable manner of the method allows application on numerous situations across data and explainability techniques.
Because there is still no perfect explainability method, a careful interpretation of explanations would be required for real-world data.
Overreliance on the explanations may mislead decisions and be vulnerable to biases.
We hope that our work will benefit GNN applications by improving the transparency of GNN models, which is highly important for critical uses requiring credibility and fairness, such as drug discovery, medical care, and social communication.

\begin{ack}
  We thank Taehee Lee for his helpful comments.
  Funding in direct support of this work was obtained from Samsung SDS.
\end{ack}

\bibliographystyle{unsrtnat}
\bibliography{references.bib}

\begin{thebibliography}{35}
\providecommand{\natexlab}[1]{#1}
\providecommand{\url}[1]{\texttt{#1}}
\expandafter\ifx\csname urlstyle\endcsname\relax
  \providecommand{\doi}[1]{doi: #1}\else
  \providecommand{\doi}{doi: \begingroup \urlstyle{rm}\Url}\fi

\bibitem[Hamilton et~al.(2017)Hamilton, Ying, and Leskovec]{graphsage}
William~L Hamilton, Rex Ying, and Jure Leskovec.
\newblock Inductive representation learning on large graphs.
\newblock In \emph{NeurIPS}, pages 1025--1035, 2017.

\bibitem[Kipf and Welling(2017)]{gcn}
Thomas~N Kipf and Max Welling.
\newblock Semi-supervised classification with graph convolutional networks.
\newblock In \emph{ICLR}, 2017.

\bibitem[Xu et~al.(2019{\natexlab{a}})Xu, Cheng, Luo, Liu, and
  Zhang]{xu2019spatio}
Dongkuan Xu, Wei Cheng, Dongsheng Luo, Xiao Liu, and Xiang Zhang.
\newblock Spatio-temporal attentive rnn for node classification in temporal
  attributed graphs.
\newblock In \emph{IJCAI}, pages 3947--3953, 2019{\natexlab{a}}.

\bibitem[Bruna et~al.(2014)Bruna, Zaremba, Szlam, and LeCun]{bruna}
Joan Bruna, Wojciech Zaremba, Arthur Szlam, and Yann LeCun.
\newblock Spectral networks and deep locally connected networks on graphs.
\newblock In \emph{ICLR}, 2014.

\bibitem[Defferrard et~al.(2016)Defferrard, Bresson, and
  Vandergheynst]{defferrard}
Micha{\"e}l Defferrard, Xavier Bresson, and Pierre Vandergheynst.
\newblock Convolutional neural networks on graphs with fast localized spectral
  filtering.
\newblock In \emph{NeurIPS}, pages 3844--3852, 2016.

\bibitem[Veli{\v{c}}kovi{\'c} et~al.(2018)Veli{\v{c}}kovi{\'c}, Cucurull,
  Casanova, Romero, Lio, and Bengio]{gat}
Petar Veli{\v{c}}kovi{\'c}, Guillem Cucurull, Arantxa Casanova, Adriana Romero,
  Pietro Lio, and Yoshua Bengio.
\newblock Graph attention networks.
\newblock In \emph{ICLR}, 2018.

\bibitem[Gilmer et~al.(2017)Gilmer, Schoenholz, Riley, Vinyals, and Dahl]{mpnn}
Justin Gilmer, Samuel~S Schoenholz, Patrick~F Riley, Oriol Vinyals, and
  George~E Dahl.
\newblock Neural message passing for quantum chemistry.
\newblock In \emph{ICML}, pages 1263--1272, 2017.

\bibitem[Xu et~al.(2019{\natexlab{b}})Xu, Hu, Leskovec, and Jegelka]{gin}
Keyulu Xu, Weihua Hu, Jure Leskovec, and Stefanie Jegelka.
\newblock How powerful are graph neural networks?
\newblock In \emph{ICLR}, 2019{\natexlab{b}}.

\bibitem[Pope et~al.(2019)Pope, Kolouri, Rostami, Martin, and
  Hoffmann]{gnngradcam}
Phillip~E Pope, Soheil Kolouri, Mohammad Rostami, Charles~E Martin, and Heiko
  Hoffmann.
\newblock Explainability methods for graph convolutional neural networks.
\newblock In \emph{CVPR}, pages 10772--10781, 2019.

\bibitem[Yuan et~al.(2021)Yuan, Yu, Gui, and Ji]{gnn_xai_taxonomy}
Hao Yuan, Haiyang Yu, Shurui Gui, and Shuiwang Ji.
\newblock Explainability in graph neural networks: A taxonomic survey.
\newblock \emph{arXiv preprint arXiv:2012.15445}, 2021.

\bibitem[Baldassarre and Azizpour(2019)]{guidedbp_LRP_gnn}
Federico Baldassarre and Hossein Azizpour.
\newblock Explainability techniques for graph convolutional networks.
\newblock In \emph{ICML Workshop on Learning and Reasoning with
  Graph-Structured Representations}, 2019.

\bibitem[Ying et~al.(2019)Ying, Bourgeois, You, Zitnik, and
  Leskovec]{gnnexplainer}
Rex Ying, Dylan Bourgeois, Jiaxuan You, Marinka Zitnik, and Jure Leskovec.
\newblock Gnnexplainer: Generating explanations for graph neural networks.
\newblock In \emph{NeurIPS}, 2019.

\bibitem[Luo et~al.(2020)Luo, Cheng, Xu, Yu, Zong, Chen, and
  Zhang]{pgexplainer}
Dongsheng Luo, Wei Cheng, Dongkuan Xu, Wenchao Yu, Bo~Zong, Haifeng Chen, and
  Xiang Zhang.
\newblock Parameterized explainer for graph neural network.
\newblock In \emph{NeurIPS}, pages 19620--19631, 2020.

\bibitem[Schlichtkrull et~al.(2020)Schlichtkrull, De~Cao, and Titov]{graphmask}
Michael~Sejr Schlichtkrull, Nicola De~Cao, and Ivan Titov.
\newblock Interpreting graph neural networks for nlp with differentiable edge
  masking.
\newblock \emph{arXiv preprint arXiv:2010.00577}, 2020.

\bibitem[Vu and Thai(2020)]{pgmexplainer}
Minh Vu and My~T Thai.
\newblock Pgm-explainer: Probabilistic graphical model explanations for graph
  neural networks.
\newblock In \emph{NeurIPS}, pages 12225--12235, 2020.

\bibitem[Duvenaud et~al.(2015)Duvenaud, Maclaurin, Aguilera-Iparraguirre,
  G{\'o}mez-Bombarelli, Hirzel, Aspuru-Guzik, and Adams]{duvenaud}
David Duvenaud, Dougal Maclaurin, Jorge Aguilera-Iparraguirre, Rafael
  G{\'o}mez-Bombarelli, Timothy Hirzel, Al{\'a}n Aspuru-Guzik, and Ryan~P
  Adams.
\newblock Convolutional networks on graphs for learning molecular fingerprints.
\newblock In \emph{NeurIPS}, pages 2224--2232, 2015.

\bibitem[Zhang and Chen(2018)]{zhang2018link}
Muhan Zhang and Yixin Chen.
\newblock Link prediction based on graph neural networks.
\newblock In \emph{NeurIPS}, pages 5171--5181, 2018.

\bibitem[Kazemi and Poole(2018)]{kazemi2018simple}
Seyed~Mehran Kazemi and David Poole.
\newblock Simple embedding for link prediction in knowledge graphs.
\newblock In \emph{NeurIPS}, pages 4289--4300, 2018.

\bibitem[Simonyan et~al.(2013)Simonyan, Vedaldi, and Zisserman]{gradient}
Karen Simonyan, Andrea Vedaldi, and Andrew Zisserman.
\newblock Deep inside convolutional networks: Visualising image classification
  models and saliency maps.
\newblock \emph{arXiv preprint arXiv:1312.6034}, 2013.

\bibitem[Zhou et~al.(2016)Zhou, Khosla, Lapedriza, Oliva, and Torralba]{cam}
Bolei Zhou, Aditya Khosla, Agata Lapedriza, Aude Oliva, and Antonio Torralba.
\newblock Learning deep features for discriminative localization.
\newblock In \emph{CVPR}, pages 2921--2929, 2016.

\bibitem[Selvaraju et~al.(2017)Selvaraju, Cogswell, Das, Vedantam, Parikh, and
  Batra]{gradcam}
Ramprasaath~R Selvaraju, Michael Cogswell, Abhishek Das, Ramakrishna Vedantam,
  Devi Parikh, and Dhruv Batra.
\newblock Grad-cam: Visual explanations from deep networks via gradient-based
  localization.
\newblock In \emph{ICCV}, pages 618--626, 2017.

\bibitem[Chattopadhay et~al.(2018)Chattopadhay, Sarkar, Howlader, and
  Balasubramanian]{gradcam_plus}
Aditya Chattopadhay, Anirban Sarkar, Prantik Howlader, and Vineeth~N
  Balasubramanian.
\newblock Grad-cam++: Generalized gradient-based visual explanations for deep
  convolutional networks.
\newblock In \emph{WACV}, pages 839--847, 2018.

\bibitem[Zintgraf et~al.(2017)Zintgraf, Cohen, Adel, and Welling]{Zint_2017}
Luisa~M Zintgraf, Taco~S Cohen, Tameem Adel, and Max Welling.
\newblock Visualizing deep neural network decisions: Prediction difference
  analysis.
\newblock In \emph{ICLR}, 2017.

\bibitem[Fong and Vedaldi(2017)]{Fong_2017}
Ruth~C Fong and Andrea Vedaldi.
\newblock Interpretable explanations of black boxes by meaningful perturbation.
\newblock In \emph{ICCV}, pages 3429--3437, 2017.

\bibitem[Fong et~al.(2019)Fong, Patrick, and Vedaldi]{Fong_2019_ICCV}
Ruth Fong, Mandela Patrick, and Andrea Vedaldi.
\newblock Understanding deep networks via extremal perturbations and smooth
  masks.
\newblock In \emph{ICCV}, pages 2950--2958, 2019.

\bibitem[Bach et~al.(2015)Bach, Binder, Montavon, Klauschen, M{\"u}ller, and
  Samek]{lrp}
Sebastian Bach, Alexander Binder, Gr{\'e}goire Montavon, Frederick Klauschen,
  Klaus-Robert M{\"u}ller, and Wojciech Samek.
\newblock On pixel-wise explanations for non-linear classifier decisions by
  layer-wise relevance propagation.
\newblock \emph{PLOS ONE}, 10\penalty0 (7):\penalty0 1--46, 2015.

\bibitem[Shrikumar et~al.(2017)Shrikumar, Greenside, and Kundaje]{deeplift}
Avanti Shrikumar, Peyton Greenside, and Anshul Kundaje.
\newblock Learning important features through propagating activation
  differences.
\newblock In \emph{ICML}, pages 3145--3153, 2017.

\bibitem[Funke et~al.(2021)Funke, Khosla, and Anand]{zorro}
Thorben Funke, Megha Khosla, and Avishek Anand.
\newblock Hard masking for explaining graph neural networks, 2021.
\newblock URL \url{https://openreview.net/forum?id=uDN8pRAdsoC}.

\bibitem[Wang et~al.(2021)Wang, Wu, Zhang, He, and Chua]{causal}
Xiang Wang, Yingxin Wu, An~Zhang, Xiangnan He, and Tat-seng Chua.
\newblock Causal screening to interpret graph neural networks, 2021.
\newblock URL \url{https://openreview.net/forum?id=nzKv5vxZfge}.

\bibitem[Schnake et~al.(2020)Schnake, Eberle, Lederer, Nakajima, Sch{\"u}tt,
  M{\"u}ller, and Montavon]{gnn_lrp}
Thomas Schnake, Oliver Eberle, Jonas Lederer, Shinichi Nakajima, Kristof~T
  Sch{\"u}tt, Klaus-Robert M{\"u}ller, and Gr{\'e}goire Montavon.
\newblock Higher-order explanations of graph neural networks via relevant
  walks.
\newblock \emph{arXiv preprint arXiv:2006.03589}, 2020.

\bibitem[Smilkov et~al.(2017)Smilkov, Thorat, Kim, Vi{\'e}gas, and
  Wattenberg]{smoothgrad}
Daniel Smilkov, Nikhil Thorat, Been Kim, Fernanda Vi{\'e}gas, and Martin
  Wattenberg.
\newblock Smoothgrad: removing noise by adding noise.
\newblock \emph{arXiv preprint arXiv:1706.03825}, 2017.

\bibitem[Oh et~al.(2021)Oh, Jung, Park, and Kim]{evet}
Youngrock Oh, Hyungsik Jung, Jeonghyung Park, and Min~Soo Kim.
\newblock Evet: Enhancing visual explanations of deep neural networks using
  image transformations.
\newblock In \emph{WACV}, pages 3579--3587, 2021.

\bibitem[Sanchez-Lengeling et~al.(2020)Sanchez-Lengeling, Wei, Lee, Reif, Wang,
  Qian, McCloskey, Colwell, and Wiltschko]{attribution}
Benjamin Sanchez-Lengeling, Jennifer Wei, Brian Lee, Emily Reif, Peter Wang,
  Wesley~Wei Qian, Kevin McCloskey, Lucy Colwell, and Alexander Wiltschko.
\newblock Evaluating attribution for graph neural networks.
\newblock In \emph{NeurIPS}, pages 5898--5910, 2020.

\bibitem[Paszke et~al.(2019)Paszke, Gross, Massa, Lerer, Bradbury, Chanan,
  Killeen, Lin, Gimelshein, Antiga, Desmaison, Kopf, Yang, DeVito, Raison,
  Tejani, Chilamkurthy, Steiner, Fang, Bai, and Chintala]{pytorch}
Adam Paszke, Sam Gross, Francisco Massa, Adam Lerer, James Bradbury, Gregory
  Chanan, Trevor Killeen, Zeming Lin, Natalia Gimelshein, Luca Antiga, Alban
  Desmaison, Andreas Kopf, Edward Yang, Zachary DeVito, Martin Raison, Alykhan
  Tejani, Sasank Chilamkurthy, Benoit Steiner, Lu~Fang, Junjie Bai, and Soumith
  Chintala.
\newblock Pytorch: An imperative style, high-performance deep learning library.
\newblock In \emph{NeurIPS}, pages 8024--8035, 2019.

\bibitem[Fey and Lenssen(2019)]{pyg}
Matthias Fey and Jan~E Lenssen.
\newblock Fast graph representation learning with pytorch geometric.
\newblock In \emph{ICLR Workshop on Representation Learning on Graphs and
  Manifolds}, 2019.

\end{thebibliography}


\section*{Checklist}
\begin{enumerate}

\item For all authors...
\begin{enumerate}
  \item Do the main claims made in the abstract and introduction accurately reflect the paper's contributions and scope?
    \answerYes{Our claims are described in Introduction section and summarized in the abstract.}
  \item Did you describe the limitations of your work?
    \answerYes{Limitations of our work in three aspects are listed in Limitation section.}
  \item Did you discuss any potential negative societal impacts of your work?
    \answerYes{Discussion on the potential negative impacts are described in Broader Impact section.}
  \item Have you read the ethics review guidelines and ensured that your paper conforms to them?
    \answerYes{Authors have read and checked the paper based on the ethics review guidelines.}
\end{enumerate}

\item If you are including theoretical results...
\begin{enumerate}
  \item Did you state the full set of assumptions of all theoretical results?
    \answerYes{We describe our conjecture on aggregating explanations in Section~\ref{motivation}.}
	\item Did you include complete proofs of all theoretical results?
    \answerNA{We do not include theoretical results require proofs.}
\end{enumerate}

\item If you ran experiments...
\begin{enumerate}
  \item Did you include the code, data, and instructions needed to reproduce the main experimental results (either in the supplemental material or as a URL)?
    \answerYes{Detailed experimental information are listed either in Section~\ref{experiments} and Appendix. Code for the experiments are to be made public in near future.}
  \item Did you specify all the training details (e.g., data splits, hyperparameters, how they were chosen)?
    \answerYes{See Section~\ref{datasets} and Appendix for details on data. Training hyperparameters are listed in Appendix. Detailed model information can be found in Section~\ref{model} and Appendix.}
	\item Did you report error bars (e.g., with respect to the random seed after running experiments multiple times)?
    \answerYes{For the standard deviations by the ten-fold experiments, see extended table in Appendix.}
	\item Did you include the total amount of compute and the type of resources used (e.g., type of GPUs, internal cluster, or cloud provider)?
    \answerYes{Hardware details for experiments are listed in Appendix.}
\end{enumerate}

\item If you are using existing assets (e.g., code, data, models) or curating/releasing new assets...
\begin{enumerate}
  \item If your work uses existing assets, did you cite the creators?
    \answerYes{We adopt synthetic datasets constructed by \citeauthor{gnnexplainer} and data splits from \citeauthor{pgexplainer}. Citations are added at their occurance (see Section~\ref{datasets} and Appendix).}
  \item Did you mention the license of the assets?
    \answerYes{We mention the license of used datasets in Appedix with corresponding citation.}
  \item Did you include any new assets either in the supplemental material or as a URL?
    \answerNA{We do not include new assets in our experiments.}
  \item Did you discuss whether and how consent was obtained from people whose data you're using/curating?
    \answerYes{The dataset is open publically via official Github repository of \citeauthor{gnnexplainer}~\cite{gnnexplainer} under Apache License 2.0. We cite the dataset and the constructors in Section~\ref{datasets} and Appendix.}
  \item Did you discuss whether the data you are using/curating contains personally identifiable information or offensive content?
    \answerNA{Data used in our research are synthetic and abstract graph datasets, which do not include personal information.}
\end{enumerate}

\item If you used crowdsourcing or conducted research with human subjects...
\begin{enumerate}
  \item Did you include the full text of instructions given to participants and screenshots, if applicable?
    \answerNA{Our work does not include experiments/research with human subjects.}
  \item Did you describe any potential participant risks, with links to Institutional Review Board (IRB) approvals, if applicable?
    \answerNA{Our work does not include experiments/research requiring IRB approvals.}
  \item Did you include the estimated hourly wage paid to participants and the total amount spent on participant compensation?
    \answerNA{Our work does not include experiments/research with human subjects.}
\end{enumerate}

\end{enumerate}


\newpage
\appendix

\section{Experimental details}
\subsection{Hardware and environment} \label{sup_environment}
Experiments are conducted on a Ubuntu 16.04 server with four RTX Titan GPUs with 24GB memory each.
Python 3.6.10 environment with CUDA version 10.1 and NVIDIA driver version 418.39 is used.
All models are implemented with Pytorch~\cite{pytorch} version 1.6.0 and Pytorch Geometric~\cite{pyg} version 1.6.1.

\subsection{Model architecture and training} \label{sup_model}
We use basic graph convolutional network (GCN)~\cite{gcn} model for all experiments.
Three-layered GCN model with node feature concatenation and fully connected layer is used.
Entire network architecture with detailed information is listed in Table~\ref{table:architecture}.
Adam optimizer with learning rate 0.001 and L2 weight decay 0.001 is adopted for all datasets, except for Tree-Grid dataset which utilized 0.002 for the weight decay.
Each model for BA-shapes, Tree-Cycles, and Tree-Grid datasets is trained for 10000 epochs, whereas models for BA-Community dataset are trained for 5000 epochs.

\begin{table}[ht]
  \caption{Model architecture and hyperparameters.}
  \vspace*{1mm}
  \label{table:architecture}
  \centering
  \begin{tabular}{llll}
    \toprule
    \textbf{Layer} & \textbf{Type}   & \textbf{Parameter}       & \textbf{Value} \\
    \midrule
    1 & Graph convolution & Input shapes    & \#Nodes$\times$\#Node features\\
      &                   & Output shapes   & \#Nodes$\times$20             \\
      &                   & Activation      & ReLU                          \\
    \midrule
    2 & Graph convolution & Input shapes    & \#Nodes$\times$20             \\
      &                   & Output shapes   & \#Nodes$\times$20             \\
      &                   & Activation      & ReLU                          \\
    \midrule
    3 & Graph convolution & Input shapes    & \#Nodes$\times$20             \\
      &                   & Output shapes   & \#Nodes$\times$20             \\
      &                   & Activation      & ReLU                          \\
    \midrule
    4 & Concatenation     & Input shapes    & Three \#Nodes$\times$20 from Layer 1, 2, 3 \\
      &                   & Output shapes   & \#Nodes$\times$60             \\
    \midrule
    5 & Fully connected   & Input shapes    & \#Nodes$\times$60             \\
      &                   & Output shapes   & \#Nodes$\times$\#Classes      \\
      &                   & Activation      & None                          \\
    \bottomrule
  \end{tabular}
\end{table}

\subsection{Dataset} \label{sup_dataset}
Datasets are taken from GNNExplainer paper~\cite{gnnexplainer} under Apache License 2.0.
Dataset split is taken from the PGExplainer code by \citeauthor{pgexplainer}~\cite{pgexplainer}, which splits train/validation/test sets by 80/10/10\%.
Averaged prediction accuracies with standard deviations for the trained models are listed in Table~\ref{table:model_acc}.
Accuracies for the datasets are considered to be high enough for explanation experiments and no further hyperparameter scan is conducted.

\begin{table}[ht]
  \caption{Prediction accuracies (AUC-ROC) for the trained models.}
  \vspace*{1mm}
  \label{table:model_acc}
  \centering
  \begin{tabular}{lcccc}
    \toprule
    \textbf{Data split}
    & \textbf{BA-Shapes} & \textbf{BA-Community} & \textbf{Tree-Cycles} & \textbf{Tree-Grid} \\
    \midrule
    Training
    & 0.966$\pm$0.029 & 0.999$\pm$0.000 & 0.944$\pm$0.005 & 0.902$\pm$0.005\\
    Validation
    & 0.973$\pm$0.021 & 0.794$\pm$0.023 & 0.951$\pm$0.005 & 0.935$\pm$0.023\\
    Test
    & 0.964$\pm$0.038 & 0.805$\pm$0.024 & 0.947$\pm$0.011 & 0.934$\pm$0.011\\
    \bottomrule
  \end{tabular}
\end{table}

\section{Supplementary results} \label{sup_results}
\subsection{Explanation accuracy}
All explanation-generating experiments are conducted in ten-fold with the ten aforementioned trained models for each dataset.
Extended accuracy table with standard deviations and p-values corresponding to the Table~\ref{table:accuracy} is shown in Table~\ref{table:accuracy_extended}.
For statistical analysis, each result is first assessed with normality test to check their distribution.
For pairs with at least one $p<0.05$ in normality test, one-sided Wilcoxon signed-rank test is used for analysis.
For the rest, one-sided Student's T-test is used.

\begin{table}[ht]
  \caption{Explanation accuracies (AUC-ROC) before and after applying $\textsc{Seen}$ with standard deviations, p-values, and their best performing $\alpha$ and $\beta$ values.}
  \vspace*{1mm}
  \label{table:accuracy_extended}
  \centering
  \begin{tabular}{lcccc}
  \toprule
  \textbf{Explainer}
  & \textbf{BA-Shapes}          & \textbf{BA-Community}
  & \textbf{Tree-Cycles}        & \textbf{Tree-Grid} \\
  \midrule
  SA
  & 0.935$\pm$0.009         & 0.625$\pm$0.020
  & 0.886$\pm$0.050         & 0.814$\pm$0.112 \\
  SA + $\textsc{Seen}$
  & 0.938$\pm$0.005         & 0.637$\pm$0.018
  & 0.900$\pm$0.035         & 0.866$\pm$0.136 \\
  p-value
  & $6.5\!\times\! 10^{-2}$ & $9.4\!\times\! 10^{-3}$
  & $1.4\!\times\! 10^{-1}$ & $5.2\!\times\! 10^{-2}$ \\
  ($\alpha,\beta$)
  & (0.5, 0.5)              & (1.0, 0.75)
  & (1.0, 0.5)              & (1.0, 0.75) \\
  \midrule
  Grad*Input
  & 0.925$\pm$0.011         & 0.614$\pm$0.015
  & 0.894$\pm$0.051         & 0.722$\pm$0.143 \\
  Grad*Input + $\textsc{Seen}$
  & 0.936$\pm$0.056         & 0.718$\pm$0.015
  & 0.934$\pm$0.033         & 0.800$\pm$0.124 \\
  p-value
  & $9.8\!\times\! 10^{-4}$ & $1.5\!\times\! 10^{-2}$
  & $9.8\!\times\! 10^{-4}$ & $2.1\!\times\! 10^{-2}$ \\
  ($\alpha,\beta$)
  & (1.0, 0.5)              & (1.0, 0.25)
  & (1.0, 0.5)              & (1.0, 0.5) \\
  \midrule
  GradCAM
  & 0.904$\pm$0.055         & 0.573$\pm$0.001
  & 0.750$\pm$0.041         & 0.757$\pm$0.029 \\
  GradCAM + $\textsc{Seen}$
  & 0.918$\pm$0.000         & 0.645$\pm$0.024
  & 0.756$\pm$0.071         & 0.785$\pm$0.060 \\
  p-value
  & $1.6\!\times\! 10^{-4}$ & $2.8\!\times\! 10^{-6}$
  & $1.4\!\times\! 10^{-1}$ & $5.2\!\times\! 10^{-2}$ \\
  ($\alpha,\beta$)
  & (0.25, 0.25)            & (1.0, 0.25)
  & (0.25, 0.5)             & (0.25, 0.5) \\
  \bottomrule
  \end{tabular}
\end{table}

\subsection{Aggregation coefficients}
\begin{figure}
  \centering
  \begin{subfigure}[b]{0.3\textwidth}
    \centering
    \includegraphics[trim={2cm 0 0cm 0}, width=\textwidth]{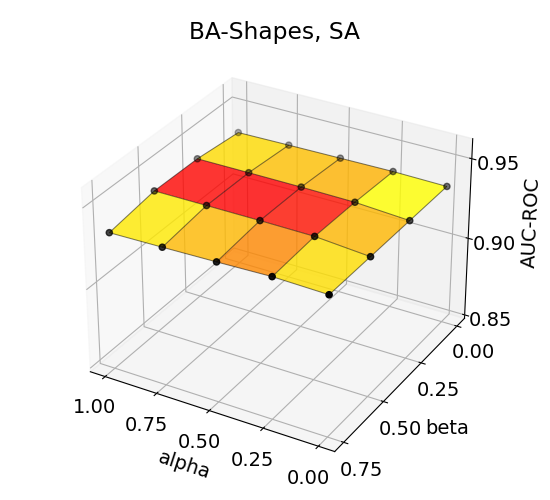}
  \end{subfigure}
  \hfill
  \begin{subfigure}[b]{0.3\textwidth}
    \centering
    \includegraphics[trim={2cm 0 0cm 0}, width=\textwidth]{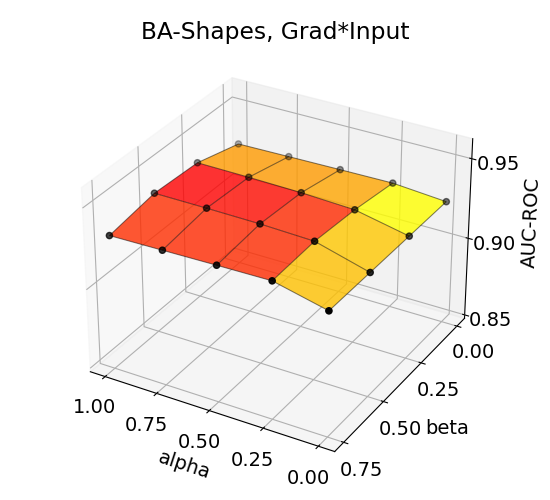}
  \end{subfigure}
  \hfill
  \begin{subfigure}[b]{0.3\textwidth}
    \centering
    \includegraphics[trim={2cm 0 0cm 0}, width=\textwidth]{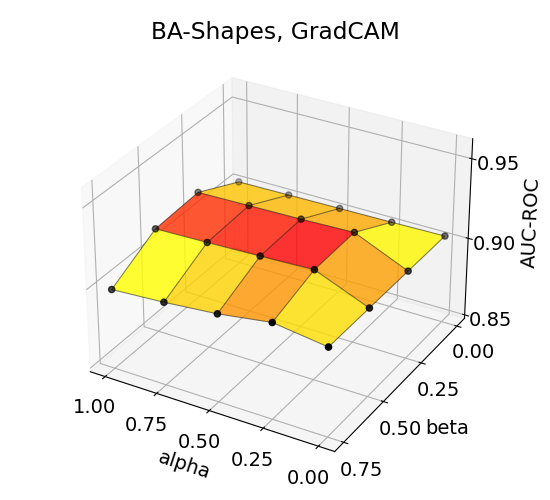}
  \end{subfigure}
  \newline

  \begin{subfigure}[b]{0.3\textwidth}
    \centering
    \includegraphics[trim={2cm 0 0cm 0}, width=\textwidth]{figures/ba-community-sa.png}
  \end{subfigure}
  \hfill
  \begin{subfigure}[b]{0.3\textwidth}
    \centering
    \includegraphics[trim={2cm 0 0cm 0}, width=\textwidth]{figures/ba-community-gradinput.png}
  \end{subfigure}
  \hfill
  \begin{subfigure}[b]{0.3\textwidth}
    \centering
    \includegraphics[trim={2cm 0 0cm 0}, width=\textwidth]{figures/ba-community-gradcam.png}
  \end{subfigure}
  \newline

  \begin{subfigure}[b]{0.3\textwidth}
    \centering
    \includegraphics[trim={2cm 0 0cm 0}, width=\textwidth]{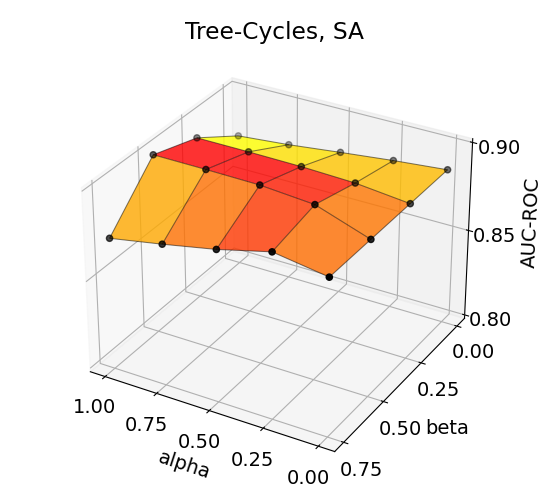}
  \end{subfigure}
  \hfill
  \begin{subfigure}[b]{0.3\textwidth}
    \centering
    \includegraphics[trim={2cm 0 0cm 0}, width=\textwidth]{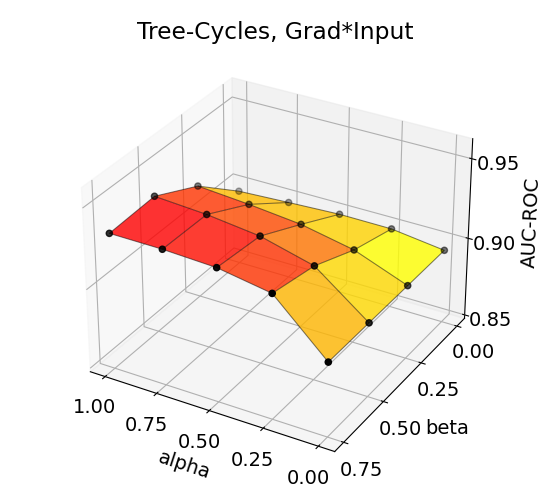}
  \end{subfigure}
  \hfill
  \begin{subfigure}[b]{0.3\textwidth}
    \centering
    \includegraphics[trim={2cm 0 0cm 0}, width=\textwidth]{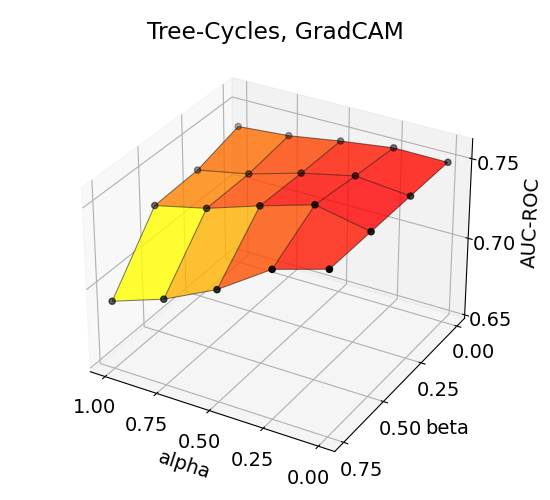}
  \end{subfigure}
  \newline

  \begin{subfigure}[b]{0.3\textwidth}
    \centering
    \includegraphics[trim={2cm 0 0cm 0}, width=\textwidth]{figures/tree-grid-sa.png}
  \end{subfigure}
  \hfill
  \begin{subfigure}[b]{0.3\textwidth}
    \centering
    \includegraphics[trim={2cm 0 0cm 0}, width=\textwidth]{figures/tree-grid-gradinput.png}
  \end{subfigure}
  \hfill
  \begin{subfigure}[b]{0.3\textwidth}
    \centering
    \includegraphics[trim={2cm 0 0cm 0}, width=\textwidth]{figures/tree-grid-gradcam.png}
  \end{subfigure}

  \caption{Effects of aggregation coefficients on explanation accuracy for BA-Shapes, BA-Community, Tree-Cycles, and Tree-Grid datasets.}
  \label{fig:coefficients_extended}
\end{figure}

Performance scan over aggregation coefficients is conducted within $\left[0, 1\right]$ for $\alpha$ and $\left[0, 1\right)$ for $\beta$ with 0.25-increament.
Note that performances for varying $\beta$ with $\alpha=0$ are identical, depicted as the right bottom line of each plot in Figure~\ref{fig:coefficients_extended}.

\end{document}